\pdfoutput=1

\documentclass[11pt]{article}

\usepackage[]{acl}

\usepackage{times}
\usepackage{latexsym}

\usepackage[T1]{fontenc}

\usepackage[utf8]{inputenc}

\usepackage{xcolor}
\usepackage{colortbl}
\usepackage{tabularx}
\usepackage{tabu}
\usepackage{booktabs}
\usepackage{multirow}

\usepackage{graphicx} 
\usepackage{float}
\usepackage{subfigure} 
\usepackage{amssymb}
\usepackage{makecell}
\usepackage{array}
\usepackage{hyperref}
\usepackage{tablefootnote}
\usepackage{xspace}

\usepackage{float}
\usepackage{amsmath}
\usepackage{bm}
\usepackage{algorithmicx}
\usepackage{algorithm}
\usepackage{algpseudocode}

\usepackage{arydshln}

\usepackage{amssymb}
\usepackage{pifont}
\newcommand{\cmark}{\ding{51}}%
\newcommand{\xmark}{\ding{55}}%
\usepackage{enumitem}

\usepackage{microtype}

\newcommand{\ours}{\textsc{VF-Eval}\xspace}
\newcommand{\method}{\textsc{RePrompt}\xspace}
\newcommand{\aivideo}{\text{AIGC video}\xspace}
\newcommand{\aivideos}{\text{AIGC videos}\xspace}
\newcommand{\eg}{\hbox{\emph{e.g.,}}\xspace}
\newcommand{\ie}{\hbox{\emph{i.e.,}}\xspace}
\newcommand{\yesno}{\textit{Yes-Or-No}\xspace}
\newcommand{\multichoice}{\textit{Multiple-choice}\xspace}
\newcommand{\openend}{\textit{Open-Ended}\xspace}

\newcommand{\cvtask}{\emph{Coherence Validation}\xspace}
\newcommand{\eatask}{\emph{Error Awareness}\xspace} 
\newcommand{\edtask}{\emph{Error Type Detection}\xspace}
\newcommand{\retask}{\emph{Reasoning Evaluation}\xspace}
\newcommand{\huggingface}{\raisebox{-1.5pt}{\includegraphics[height=1.05em]{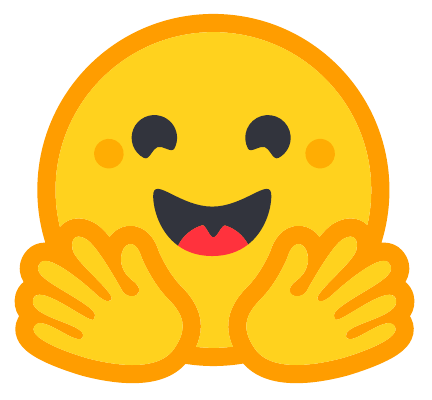}}\xspace}
\newcommand{\github}{\raisebox{-1.5pt}{\includegraphics[height=1.05em]{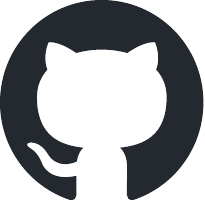}}\xspace}
\newcommand{\CV}{\emph{Coherence Validation}\xspace}
\newcommand{\EA}{\emph{Error Awareness}\xspace} 
\newcommand{\ED}{\emph{Error Type Detection}\xspace}
\newcommand{\RE}{\emph{Reasoning Evaluation}\xspace}

\newcommand{\nquestion}{9,740\xspace}
\newcommand{\nmodel}{13\xspace}

\definecolor{Gray}{gray}{0.95}
\newcolumntype{a}{>{\columncolor{Gray}}c}

\title{\ours: Evaluating Multimodal LLMs for Generating \\
Feedback on AIGC Videos}

\author{
Tingyu Song$^\clubsuit$ \quad Tongyan Hu$^\diamondsuit$ \quad Guo Gan$^\heartsuit$ \quad Yilun Zhao$^\spadesuit$ \vspace{10pt}\\
 {\small\fontsize{11.2pt}{12pt}\selectfont $^\clubsuit$ School of Advanced Interdisciplinary Sciences, University of Chinese Academy of Sciences} \quad \\
{\small\fontsize{11.2pt}{12pt}\selectfont $^\diamondsuit$ National University of Singapore } \quad
{\small\fontsize{11.2pt}{12pt}\selectfont $^\heartsuit$ Zhejiang University} \quad
 {\small\fontsize{11.2pt}{12pt}\selectfont $^\spadesuit$ Yale University} \\
}

\begin{document}
\maketitle

\begin{abstract}
Multimodal large language models (MLLMs) have been widely studied for video question answering recently. However, most existing assessments focus on natural videos, overlooking synthetic videos, such as AI-generated content (AIGC). Meanwhile, some works in video generation rely on MLLMs to evaluate the quality of generated videos, but the capabilities of MLLMs on interpreting AIGC videos remain largely underexplored. 
To address this, we propose a new benchmark, \ours, which introduces four tasks—coherence validation, error awareness, error type detection, and reasoning evaluation—to comprehensively evaluate the abilities of MLLMs on AIGC videos. We evaluate \nmodel frontier MLLMs on \ours and find that even the best-performing model, GPT-4.1, struggles to achieve consistently good performance across all tasks. This highlights the challenging nature of our benchmark.
Additionally, to investigate the practical applications of \ours in improving video generation, we conduct an experiment, \method, demonstrating that aligning MLLMs more closely with human feedback can benefit video generation.

\begin{small}
\begin{center}
\begin{tabular}{cll}
\huggingface & \textbf{Data} & \href{https://huggingface.co/datasets/songtingyu/vf-eval} {\path{songtingyu/VF-Eval}}\\
\github & \textbf{Code} & \href{https://github.com/SighingSnow/VF-Eval}{\path{SighingSnow/VF-Eval}}\\
\end{tabular}
\end{center} 
\end{small}
\end{abstract}

\begin{figure*}[!t]
    \centering
    \includegraphics[width=\textwidth]{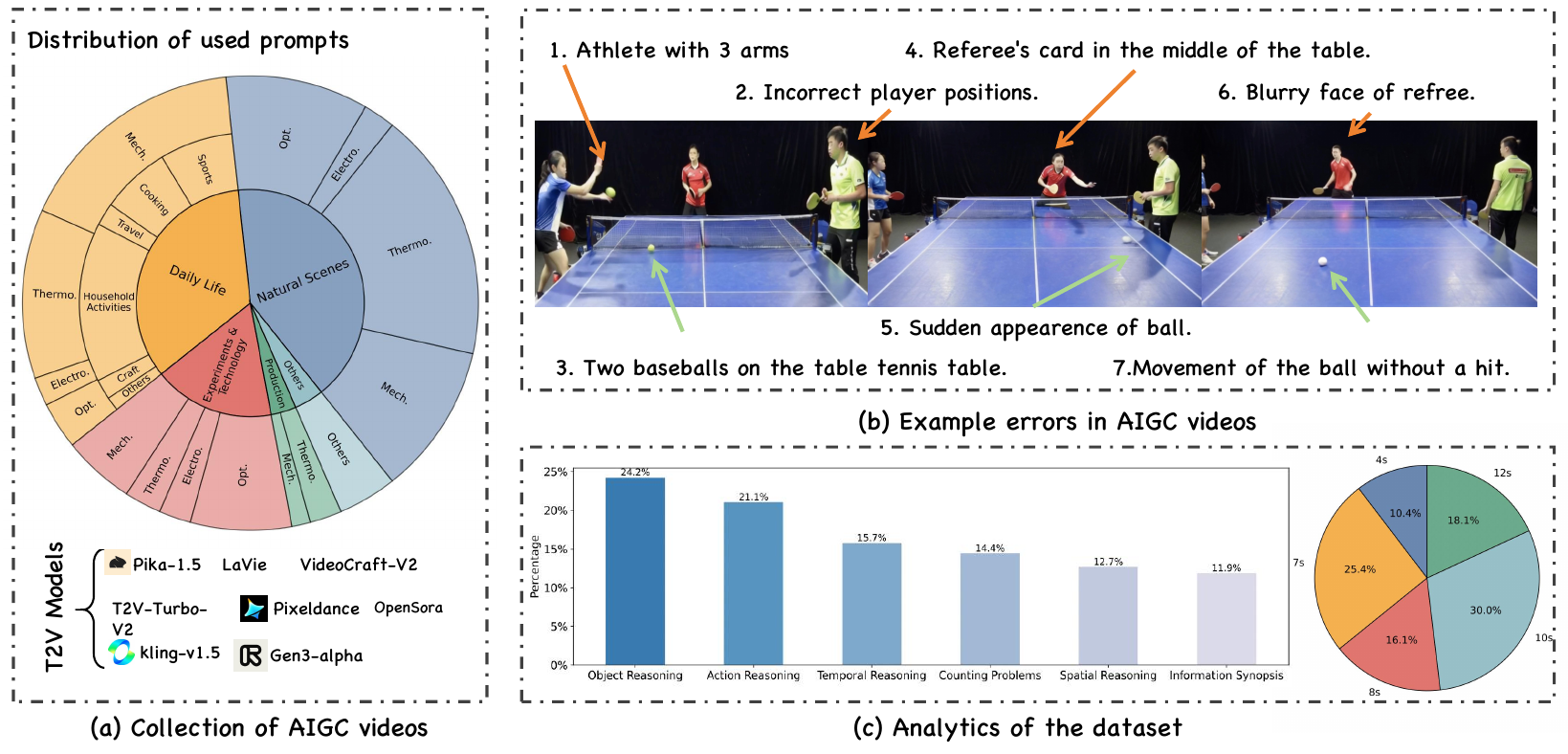}
    \caption{Overview of our research: (a) Collection of AIGC videos: We compile a diverse set of video generation prompts to instruct both proprietary and open-source T2V models for generating AIGC videos. (b) Illustration of errors occurring within the same AIGC video. (c) Analytics of the dataset: \ours covering a diverse range of reasoning tasks. And it contains AIGC videos with durations between 4 to 12 seconds, reflecting the typical output length of current T2V models.}
    \label{fig: overview}
\end{figure*}
\section{Introduction}

Multimodal Large Language Models (MLLMs) are powerful tools that process and integrate information across visual and textual domains~\cite{geminiteam2024gemini,cogvlm, qwen2-vl, videochat, llava-onevision}. While their primary applications have historically included tasks such as natural language processing~\cite{lyu2023macaw, liang2024comprehensive}, image captioning~\cite{liu2024tempcompass, bucciarelli2024personalizingmultimodallargelanguage}, and video analysis~\cite{fu2024videomme, ren2024timechat}, they are now increasingly being utilized in the domain of video generation.
In the context of video generation, MLLMs are not only applied for video quality assessment~\cite{meng2024worldsimulatorcraftingphysical, wu2024qbench}, but also play a critical role in enhancing the video creation process~\cite{kondratyuk2023videopoet, wang2024gpt4videounifiedmultimodallarge}. 
By providing feedback on generated videos—ranging from content quality to more intricate aspects like visual coherence and temporal consistency—MLLMs are applied to help improve the quality of \aivideo generation~\cite{vlrewardbench, prm-picture}.

AIGC videos present new challenges for visual understanding~\cite{aigc-challenge, aigc-challenge-2}, including synthetic textures, dynamic lighting effects, and algorithmically generated characters that significantly from those found in traditional video content. 
These distinctive characteristics complicate accurate interpretation by MLLMs, thereby reducing the reliability and effectiveness of their feedback.
Despite these challenges, existing research on MLLMs providing feedback (\ie quality assessment) on \aivideos has its limitations. 
In video quality assessment, MLLMs are often tasked with providing implicit scores~\cite{q-align, lmm-vqa}, which can be imprecise and fail to capture the full range of video quality nuances. 
While some studies focus on generating natural language feedback to assess video quality~\cite{wu2024qbench, edit-vqa}, the feedback may lack precision, especially when applied to \aivideos, whose characteristics differ significantly from traditional natural videos. 

To bridge this gap, we propose a new benchmark named \ours, designed to evaluate the capabilities of MLLMs to generate reliable feedback for \aivideos. This benchmark focuses on assessing key aspects such as alignment with expected outcomes, feedback quality, and commonsense reasoning. Specifically, we propose four tasks to systematically measure the MLLM's feedback generation capabilities:
(1) \cvtask: Detecting misalignment between the \aivideo and its generation prompt, and providing a more appropriate video generation prompt. 
(2) \eatask: Identifying errors in a video set that includes both natural and \aivideos.
(3) \edtask: Identifying possible errors within \aivideos.
(4) \retask: Demonstrating fine-grained reasoning ability over \aivideo. 
We also incorporate six reasoning tasks in \retask: spatial and temporal reasoning, action and object reasoning, counting problems, and information synopsis. 

Our experimental results across \nmodel frontier MLLMs highlight three key findings: (1) MLLMs struggle with \aivideo tasks due to the unique characteristics of \aivideos. (2) MLLMs can be utilized alongside auxiliary methods to provide more accurate feedback on corresponding tasks. (3) Open-source models demonstrate competitive performance compared to proprietary models and can be further improved for relevant tasks.

To demonstrate the potential of MLLM feedback, we conduct an experiment, \method, that compares MLLM with humans in providing video generation prompts. Through the experiments, we find that the quality and coherence of AI-generated content can be enhanced potentially by aligning MLLM with human preferences.

We conclude our contribution as follows:
\begin{itemize} [leftmargin=*]
\itemsep0em 
\item We introduce \ours, a benchmark designed to evaluate the reasoning abilities of MLLMs on interpreting \aivideos, with the goal of advancing AIGC video generation processes.
\item We conduct extensive experiments with state-of-the-art MLLMs and perform fine-grained evaluations of their reasoning capabilities across 6 critical tasks, highlighting the broader implications of our findings for future model development.
\item We conduct \method experiment, comparing MLLM and human feedback in the context of video generation prompts. Our results demonstrate that aligning MLLM feedback with human preferences can potentially enhance the quality and coherence of AI-generated videos. 
\end{itemize}

\section{Related works}

\subsection{Video Understanding Benchmark} 
Recently, numerous MLLMs~\cite{geminiteam2024gemini,  cogvlm, chatglm, llava-onevision, deepseekvl2, qwen25vl} have been introduced, showcasing strong competencies in handling multimodal inputs and delivering appropriate responses.
Various benchmarks~\cite{videovista, mmbench, rextime, mvbench, mlvu-ac, zhao-etal-2025-multimodal, zhao2025mmvu} have been proposed to test these models across various scenarios, as presented in \autoref{tab: dataset-compare}.
Moreover, some studies try to evaluate the MLLM's capacity to grasp commonsense and physical knowledge as presented in videos~\cite{acquired, physbench, visfactor}. 
However, existing datasets are typically based on natural videos, leaving the reasoning capabilities on AIGC videos underexplored. 
Therefore, \ours is proposed to assess MLLM comprehension skills on \aivideos through four distinct tasks. 

\subsection{Evaluation of AIGC Video Generation}
As video generation becomes increasingly popular, various methods have emerged to evaluate its quality. Traditional video quality assessment techniques for user-generated content videos ~\cite{Tu_2021, ging2024openendedvqabenchmarkingvisionlanguage} and \aivideos~\cite{vbench,fan2024aigcbench, fetv} heavily utilize computer vision methods, offering quantitative scores that partially capture the perceived quality of videos. However, these scores fall short of identifying areas of divergence from human preferences or areas needing enhancement. 
Meanwhile, video quality assessment methods utilizing MLLMs are better aligned with human perceptions by integrating reasoning abilities into their evaluation processes. 
Recent studies~\cite{wu2024qbench, meng2024worldsimulatorcraftingphysical, wang2024gpt4videounifiedmultimodallarge} have explored the use of MLLMs to deliver more interpretable assessments of video quality. While these efforts primarily emphasize overall quality evaluation, our work shifts the focus toward benchmarking the reasoning abilities of MLLMs.  
Specifically, we introduce a benchmark comprising four diverse tasks designed to evaluate MLLMs' capacity to provide detailed feedback on \aivideos, including their effectiveness in diagnosing quality issues and identifying specific errors.
\begin{figure*}[ht]
    \centering
    \includegraphics[width=0.98\textwidth]{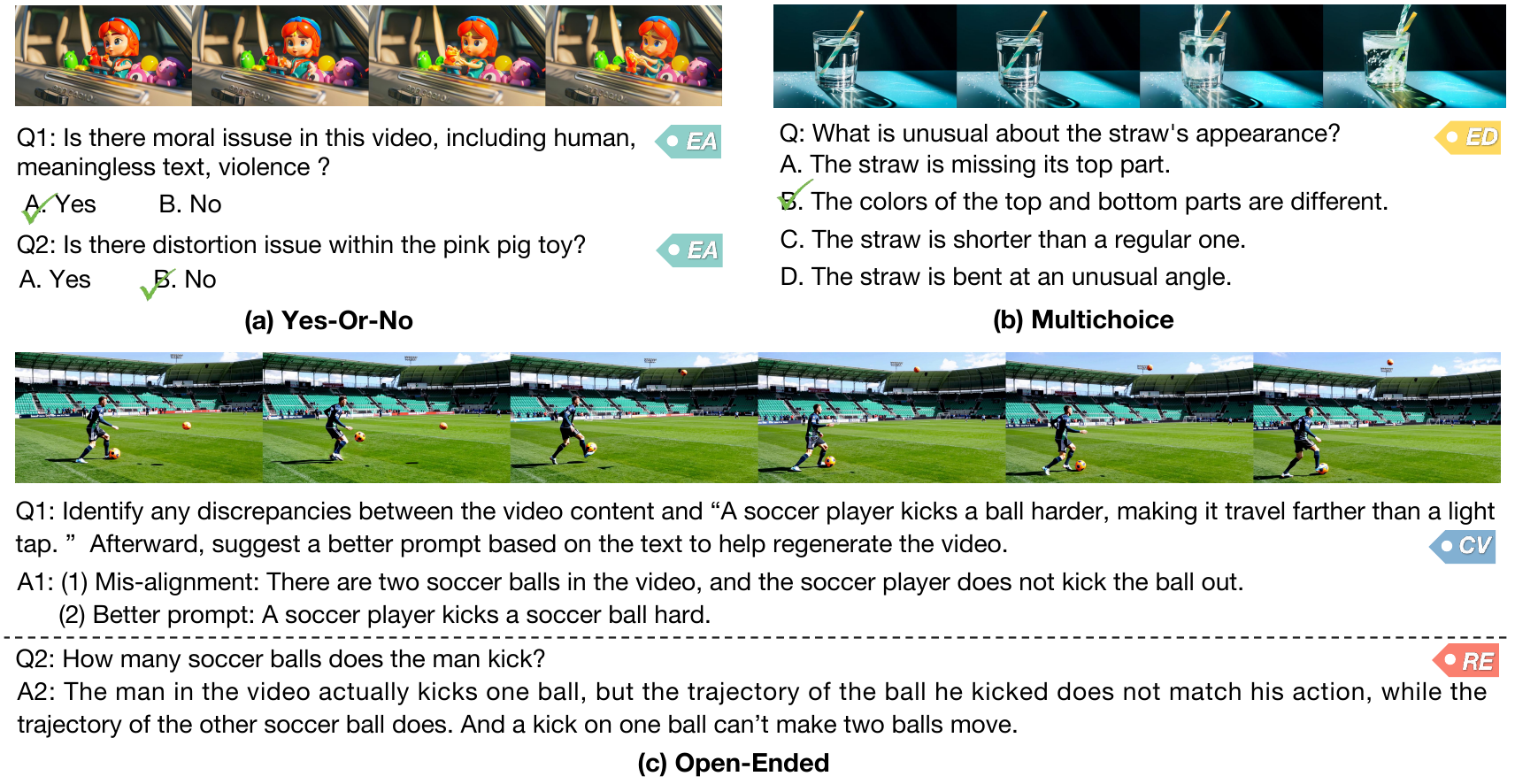}
    \caption{Illustration of four proposed tasks and the corresponding question types in the \ours benchmark.  Detailed examples for each reasoning task are provided in Appendix~\ref{appendix-reasoning-examples}. }
    \label{fig: data-example}
\end{figure*}

\begin{table}[!t]
\centering
\renewcommand{\arraystretch}{1.05}
\scalebox{0.66}{ \begin{tabular}{lp{2.3cm}ccc}
\toprule[.1em]
\multirow{2}{*}{\textbf{Benchmarks}} & \multirow{2}{*}{\textbf{QA Types}} & \multicolumn{3}{c}{\textbf{Tasks}} \\
\cmidrule(lr){3-5}
 & & \textbf{CV} & \textbf{ER} & \textbf{RE}\\
\midrule
\multicolumn{5}{c}{\textbf{Natural Videos}} \\
MVBench \cite{mvbench} & MC &  &  & \checkmark \\
AutoEval-Video\cite{chen2023autoeval} & Open &  &  & \checkmark \\
Video-Bench \cite{ning2023video} & MC &  &  &  \checkmark\\
TempCompass \cite{liu2024tempcompass} & T/F, MC, Open &  &  & \checkmark\\
TOMATO~\cite{shangguan2024tomato} & MC &  &  & \checkmark\\
Video-MME \cite{fu2024videomme} & MC &  &  & \checkmark \\
VideoVista \cite{videovista}& MC &  &  & \checkmark \\
SOK-Bench \cite{wang2024sok}& MC &  &  & \checkmark \\
MLVU \cite{zhou2024mlvu}& MC, Open &  &  & \checkmark \\
MMWorld \cite{he2024mmworld}& MC, Open &  &  & \checkmark \\
MMVU \cite{zhao2025mmvu}& MC, Open &  &  & \checkmark \\
VSI-Bench~\cite{yang2024thinking} & MC, Open &  &  & \checkmark \\
\midrule
\multicolumn{5}{c}{\textbf{Synthetic Videos}} \\
EditVid-QA \cite{edit-vqa} & Open &  &  & \checkmark\\
QBench \cite{wu2024qbench}& T/F, MC, Open &  & \checkmark &\\
\midrule
\textbf{\ours (ours)} & T/F, MC, Open & \checkmark & \checkmark & \checkmark \\
\bottomrule[.1em]
\end{tabular}
 }
\caption{Comparisons between \ours and existing video understanding benchmarks. \textbf{CV} denotes \CV, \textbf{ER} denotes \emph{error reasoning} including \EA and \ED, and \textbf{RE} denotes \RE. ``T/F'' denotes the \yesno questions, ``MC'' denotes the \multichoice questions, ``Open'' denotes the \openend questions.}
\label{tab: dataset-compare}
\end{table}

\section{\ours Benchmark}
This section first introduces the four tasks in \ours, followed by a detailed explanation of the dataset construction process for each task and an analysis of the corresponding data statistics.

\subsection{Task Formulation}\label{sec:task-formulation}
\ours includes four tasks: \CV, \EA, \ED, and \RE, each evaluated through specific question types, as shown in ~\autoref{fig: data-example}. \CV evaluates MLLMs in two key areas: assessing the alignment between the generated prompt and the corresponding video content, and determining how well MLLMs can generate prompts that align with human expectations for subsequent video generation. 
\EA and \ED focus on error detection in \aivideo, with \EA targeting the identification of general errors in videos, while \ED provides a more granular evaluation of MLLM capabilities across multiple dimensions. Recognizing that misalignment QA and error detection alone may not comprehensively evaluate MLLM performance, we introduce \RE to measure MLLMs' general reasoning abilities in the context of \aivideo. The tasks are formally defined as follows:

\paragraph{Task 1: \cvtask (CV). }   
\CV aims to verify the alignment between prompts and their corresponding \aivideo. \CV primarily relies on \openend questions to verify the alignment between prompts and their corresponding \aivideos.
The MLLM is required to compare the alignment between the video and the generation prompt, and to provide an improved prompt for generation.
Given a video \(v\), a human answer \(y\), and the answer from the MLLM \(\hat{y}\), \CV uses an LLM (i.e., GPT-4.1-mini) to rate the generated answer \(\hat{y}\) against the correct answer \(y\).
The final score is calculated as:  
\begin{equation}
\text{Score}_{CV} = \frac{1}{N} \sum_{i=1}^N LLM(y_i, \hat{y}_i),
\label{eq:cv_score}
\end{equation}

\paragraph{Task 2: \eatask (EA). }  
\EA aims to detect whether there are errors in the \aivideo. This task is primarily evaluated using \yesno questions. Given an \aivideo \(v\) and a question \(q\), the model is required to predict a label \(y\) indicating whether \(v\) contains errors. The final score for \EA is defined as:  
\begin{equation}
\text{Score}_{EA} = \frac{1}{N} \sum_{i=1}^N \mathbb{I}(y_i = \hat{y}_i),
\label{eq:ea_score}
\end{equation}

\paragraph{Task 3: \edtask (ED). }  
\ED intends to identify all the errors present in the \aivideo. This task is mainly evaluated using \multichoice questions, 
We evaluate through  the overall success rate. The score is calculated as:  
\begin{equation}
\text{Score}_{ED} = \frac{1}{N} \sum_{i=1}^N \mathbb{I}(y_i = \hat{y}_i),
\label{eq:ed_score}
\end{equation}
where \(y_i\) is the correct choice, and \(\hat{y}_i\) is the choice predicted by the MLLM.

\paragraph{Task 4: \retask (RE). }   
\RE is dedicated to evaluating the reasoning ability of MLLMs on complex questions.
As shown in ~\autoref{fig: overview}, we have six sub-tasks: spatial and temporal reasoning, action and object reasoning, counting problems, and information synopsis.  
We provide the definition and illustrative examples for each task in Appendix \ref{appendix-reasoning-examples}. 
And the evaluation is primarily realized through \openend questions. 
Given a question \(q\) and an \aivideo \(v\), \RE used an LLM (\ie GPT-4.1-mini) to evaluate the MLLM's response \(\hat{y}\) against the human-provided answer \(y\). The final score for \RE is computed as:  
\begin{equation}
\text{Score}_{RE} = \frac{1}{N} \sum_{i=1}^N LLM(y_i, \hat{y}_i),
\label{eq:re_score}
\end{equation}
where \(N\) represents the number of evaluations.

\subsection{Dataset Construction Guidelines}
To ensure the high quality of our dataset, \ours adheres to the following collection guidelines: 
(1) \textbf{Wide Scenarios Coverage}: To realize this, we generate videos using 1000 prompts generated by LLM (\ie GPT-4o). As shown in ~\autoref{fig: overview}(a), the prompts are validated by human experts and presented in the Appendix. 
Additionally, we collect other videos from existing datasets. 
(2) \textbf{Knowledge Intensive}: We carefully craft the options in \multichoice and \openend question, incorporating commonsense and physical knowledge (\eg mechanics, light, material). This approach requires MLLMs to leverage their expertise and analytical skills to address the related issues. 
(3) \textbf{Reasoning Ability}: We carefully design the \multichoice and \openend problems. For the \multichoice questions, we employ MLLM (\ie GPT-4o) to create distracting options, subsequently verified by human reviewers and combined with the accurate responses. Regarding \openend questions, we evaluate MLLM's capacities for spatial, temporal, action, and object reasoning, as well as counting and information synopsis, using \aivideos.

\subsection{Dataset Construction}

To benchmark the reasoning abilities of MLLMs on \aivideos, we collect a large-scale \aivideo dataset that ensures a wide range of diversity in video content and scenarios. 
We design video generation prompts to cover various daily scenarios, providing a comprehensive foundation for evaluating MLLMs’ reasoning capabilities. To enhance diversity, we use both proprietary and open-source video generation models. For proprietary models, we select Pika, Kling, Pixeldance, and Gen-3. while for open-source models, we include videos generated by T2V-turbo-v2~\cite{t2v-turbo-v2}. In addition to these generated videos, we enrich our dataset by collecting \aivideos from existing datasets, specifically, from Lavie~\cite{lavie} and OpenSora~\cite{opensora} in the Videophy~\cite{bansal2024videophy} train split.


\paragraph{\multichoice Question Annotation. }   
\multichoice questions are intended to benchmark the \ED task. They are constructed through a pipeline involving both human annotators and MLLMs. Initially, human annotators identify errors across three dimensions: (1) \emph{Video Quality}, which includes aspects such as temporal-spatial coherence, visual appeal, and camera work; (2) \emph{Commonsense and Physical Violations}, which encompass logical inconsistencies, mechanical flaws, lighting issues, and other abnormalities; and (3) \emph{Morality}, which addresses concerns like fear inducement, human portrayal, textual content, and graphic violence. After the human annotators provide answers, MLLMs are tasked with generating distracting options. For this, the input includes both the videos and the question-answer pairs. The MLLM-generated misleading answers are reviewed by human annotators and used to complement the original question-answer pairs. Once the options are finalized, \multichoice questions are constructed with fine-grained granularity. For example, ``Given the video, select the choice that influences the video quality'' or ``Select the choices that reflect the abnormal behavior of the bicycle in the video.''

\paragraph{\yesno Question Annotation. }  
The \yesno questions are primarily designed for the \EA task, prompting MLLMs to make binary judgments. We utilize LLM(\ie GPT-4o) to convert \multichoice questions into \yesno questions. 
All questions in our Yes-or-No task are designed with “Yes” as the correct answer. This intentional setup allows us to investigate whether MLLMs exhibit bias toward perceiving videos as normal. 
The question is like ``Check whether this video contains any commonsense violations.''

\paragraph{\openend Question Annotation.}  
\openend questions cover both the \CV and \RE tasks. For the \CV task, annotators are provided with the \aivideo and the prompt used to generate the video. They are instructed to provide two answers: the misalignment between the video and the prompt, and a revised prompt that they believe would generate a better video. Questions for this task include ``Given a prompt and the video generated by it, could you provide a better prompt to generate a more accurate video?'' or ``Could you point out the misalignment between the video and the given prompt?'' For the \RE task, human annotators construct questions across several reasoning categories, including spatial reasoning, temporal reasoning, action reasoning, object reasoning, counting problems, and information synopsis tasks. For example, a question for the spatial reasoning type could be ``Please specify the relationship between the planet and the astronauts.''

\subsection{\ours Data Analysis}

\paragraph{Dataset Statistics. } We present the statistics of \ours in ~\autoref{tab: sec3-data-analysis}. 
\ours includes a total of \nquestion question-answer pairs, including 1,826 \yesno, 5,932 \multichoice, 1,982 \openend questions. And we split them into the test and validation sets. And we provide longer videos in \ours compared to existing works. 

\begin{table}[!t]
\centering
\small
\begin{tabular}{lr}
    \toprule[.1em]
     \textbf{Statistics} & \textbf{Value}  \\
    \midrule
    \emph{Dataset Split} \\ 
    \quad Test split &  6,822 \\
    \quad Validation split & 2,918 \\
    \quad Total & \nquestion \\
    \midrule
    \emph{Question Type} & \\
    \quad \yesno & 1,836 \\
    \quad \multichoice & 5,932 \\
    \quad \openend & 1,982 \\
    \midrule
    \emph{Length} & \\
    \quad Video Length (avg. / max) & 8.98 / 12 \\  
    \quad Question Length (avg. / max) & 35.25 / 119 \\
    \bottomrule[.1em]
\end{tabular}
\caption{Statistics of \ours, including the number of questions across different data splits and question types, as well as the average and maximum video length(seconds) and question length(words).}
\label{tab: sec3-data-analysis}
\end{table}

\paragraph{Human Validation.}
To guarantee the quality of \ours, we introduce a human validation process. Expert validation is introduced in the following process:  (1) Data Construction Stage: When construction, a second annotator is introduced to judge the first annotator's annotation and provide agreement. The second annotator is responsible for choice validation in \multichoice questions and checks the \openend question-answer pair quality; (2) Post Validation Stage: After \ours is constructed, we select 3 annotators with top inter-agreement scores to check all the question-answer pairs. After this validation, 2,395 question-answer pairs are corrected. And from the low percentage of revisions, we can guarantee the high quality of \ours. We also provide the details of human validation in the Appendix~\ref{appendix-annot}, including annotation UI, annotators' identity and tasks, and the inter-annotator agreement. 

\begin{table*}[htbp]
\centering
\small
\begin{tabular}{lccccccccc}
\toprule[.1em]
\renewcommand{\arraystretch}{1.1}
\multirow{2}{*}{Model} & \multirow{2}{*}{\shortstack{Coherence\\[0.8ex]Validation}} & \multicolumn{2}{c}{Error Awareness} & \multicolumn{4}{c}{Error Type Detection} &  \multirow{2}{*}{\shortstack{Reason.\\[0.8ex]Eval}} & \multirow{2}{*}{Overall}   \\
 \cmidrule(lr){3-4}  \cmidrule(lr){5-8}
 & & Quality & CP & Quality & CP & Morality & Object &  &  \\
\midrule
Random Guess & - & 50.0 & 50.0 & 25.0 & 25.0 & 25.0 & 25.0 &  - & -\\
Human & 81.9 & 84.3 & 84.2 & 86.9 & 78.3 & 93.2 & 82.1 & 70.1 & 84.4 \\
\midrule
GPT-4.1 & \textbf{66.3} & 39.7 & 24.0 & \textbf{56.0} & 49.6 & \textbf{59.4} & \textbf{75.2} & \textbf{42.1} & \textbf{51.6} \\
GPT-4.1-mini & 55.3 & 25.5 & 10.9 & 46.6 & \textbf{56.9} & 52.8 & \underline{69.0} & \underline{37.4} & \underline{44.3} \\
Gemini-2.0-Flash & 58.1 & \underline{54.7} & 1.8 & 27.9 & 21.1 & 27.4 & 46.4 & 26.0 & 32.9 \\
\noalign{\vskip 0.5ex}\hdashline\noalign{\vskip 0.5ex}
InternVL3-38B & 52.9 & 34.7 & 5.0 & \underline{49.4} & \underline{53.9} & 49.1 & 67.9 & 36.2 & 43.6 \\
InternVL3-8B & 48.6 & 42.0 & 6.1 & 35.9 & 45.7 & \underline{53.8} & 63.9 & 35.8 & 41.5 \\
Mistral-Small-3.1-24B & 57.6 & 28.7 & 12.0 & 45.5 & 36.0 & 47.2 & 53.3 & 36.3 & 39.6 \\
Llama-3.2-11B-Vision & 17.8 & \textbf{55.8} & \textbf{39.3} & 32.1 & 41.6 & 35.8 & 53.8 & 23.2 & 37.4 \\
Qwen2.5-VL-72B & \underline{59.8} & 22.9 & 8.6 & 31.0 & 34.7 & 38.7 & 55.1 & 35.6 & 35.8 \\
VideoLLaMA3-7B & 35.4 & 10.2 & 5.8 & 31.4 & 40.3 & 43.4 & 54.8 & 33.5 & 31.8 \\
Qwen2.5-VL-7B & 51.5 & 23.4 & 6.1 & 23.8 & 27.1 & 28.3 & 47.6 & 35.3 & 30.4 \\
Llama3-LLaVA-Next-8B & 48.7 & 34.5 & 15.5 & 22.3 & 25.2 & 28.3 & 36.7 & 27.1 & 29.8 \\
LLaVA-NeXT-Video-7B & 28.0 & 46.6 & 25.9 & 23.6 & 28.2 & 24.5 & 34.8 & 22.5 & 29.3 \\
Phi-3.5-Vision & 38.5 & 4.2 & 0.5 & 23.2 & 27.3 & 37.7 & 52.7 & 26.0 & 26.3 \\
\bottomrule[.1em]
\end{tabular}
\caption{Model performance (\ie accuracy \%) on \ours. \textbf{CP} denotes ``Commonsense and physics''.} 
\label{tab: results}
\end{table*}

\section{Experiments}
This section outlines the experimental setup and summarizes the key findings.

\subsection{Experiment Settings}
We evaluate a wide range of MLLMs on \ours. Specifically, we evaluate seven series of \textbf{open-source models}, including 
InternVL3~\cite{internvl3}, 
LLava-NeXT~\cite{liu2024llavanext}, 
LLaVA-NeXT-Video~\cite{llavanext-video}, 
Llama-3.2-Vision~\cite{llama3}, 
VideoLlaMA3~\cite{videollama3},
Phi-3.5-Vision~\cite{phi-3}, 
Qwen2.5-VL~\cite{qwen2-vl}, 
Mistral-Small-3.1~\cite{mistral3},
We also evaluate two series of \textbf{proprietary models}, including GPT-4.1~\cite{OpenAI2023GPT4TR}, GPT-4.1-mini and Gemini-Flash-2.0~\cite{gemini-2}. 
For models without native video support, we provide visual input according to the maximum number of images that fit within the model’s context window. 
\S\ref{appendix-mllm-conf} details the settings for different models.
We evaluate these models using the Chain-of-Thought (CoT) technique, as illustrated in Appendix~\ref{appendix-prompt}. 
\begin{figure}[!t]
    \centering
    \includegraphics[width=0.47\textwidth]{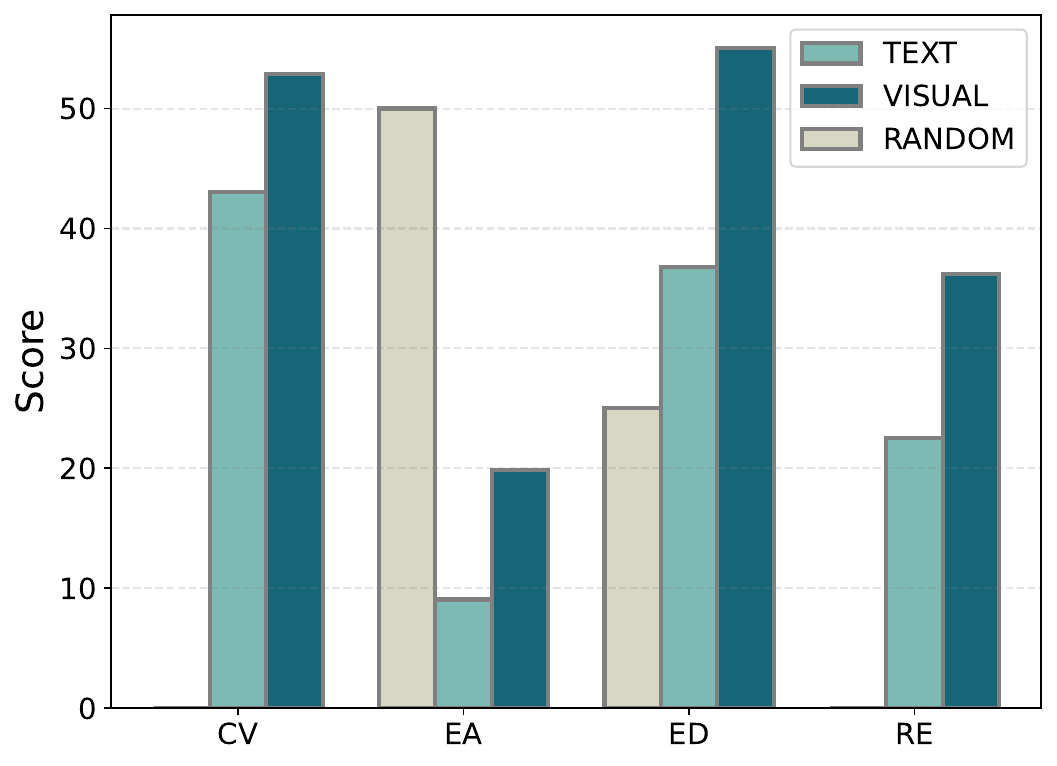}
    \caption{Performance Comparison of InternVL3-38B. 
    } 
    \label{fig: textonly-perform}
\end{figure}

\subsection{Main Findings}

\paragraph{Dataset Quality.} 
As illustrated in ~\autoref{fig: textonly-perform}, 
the absence of visual data leads to a significant decrease in performance, particularly in tasks \CV and \RE. 
We acknowledge the potential biases present in the \EA and \ED tasks, where questions might contain clear clues to the correct answer. 
However, the performance on \EA is worse than random guessing, indicating that MLLMs tend to perceive the video as normal. 
Moreover, as shown in ~\autoref{tab: results}, GPT-4.1, the model with the best overall performance, is still far from human behavior in each sub-task. 
This underscores the importance of our dataset, as it helps reveal these gaps in MLLM performance. 
Given these gaps, directly using feedback from an MLLM in video quality assessment tasks, or any other task, may lead to inaccurate results. 

\paragraph{Overall Performance.} 
From ~\autoref{tab: results}, we can deduce that the scaling law applies to this scenario. 
Additionally, we observe that there is a large performance gap between the best-performing proprietary model and the best-performing open-source models. 
This highlights the potential for open-source models to achieve competitive performance with proprietary models through fine-tuning methods. 
However, at present, MLLMs do not perform relatively well on the corresponding tasks. Other approaches, such as computer vision methods, should be incorporated as auxiliary tools to improve feedback generation.

\paragraph{Task-Specific Performance Variations }  
Performance varies across different tasks and models. We summarize our findings as follows: 
(1) In the \EA task, MLLMs perform better in the ``Quality'' aspect than in the ``CP''. 
From our analysis, this is because MLLMs lack knowledge of video quality assessment. The better performance in the ``Quality'' aspect of the \EA task could be attributed to LLM cannot detect the subtle violation in ``CP''.  
(2) In the \ED task, MLLMs perform worse than expected. As current MLLMs are aligned with human preferences, they may still fail to distinguish the morality violation in \aivideos. 
This highlights MLLMs' limitations in effectively utilizing visual inputs and their inadequate consideration of moralities depicted in videos. 

\paragraph{Challenges on \aivideos.} 
MLLMs lack knowledge of video generation. 
In task \CV, although they can largely identify the misalignment between the prompt and the video, MLLMs cannot always generate a better prompt for video generation. The prompts they provide are often simple expansions of the original prompt. Furthermore, as shown in ~\autoref{tab: results}, MLLMs may easily fail on reasoning tasks involving \aivideos. 
Due to issues such as blurriness, sudden appearances, and disappearances in \aivideos, MLLMs often struggle to capture all the details. Additionally, the unusual structure and abrupt changes in the videos may contradict MLLMs’ commonsense knowledge, resulting in worse performance on tasks \CV and \RE. Specifically, we provide a detailed analysis of the error cases in \S\ref{sec5-error-analysis}.

\section{Analysis}
We next present an in-depth analysis of MLLM reasoning capabilities and provide a detailed error analysis. Additionally, we explore methods to enhance MLLM feedback generation through fine-tuning techniques.
\begin{figure}[!t]
    \centering
    \includegraphics[width=0.47\textwidth]{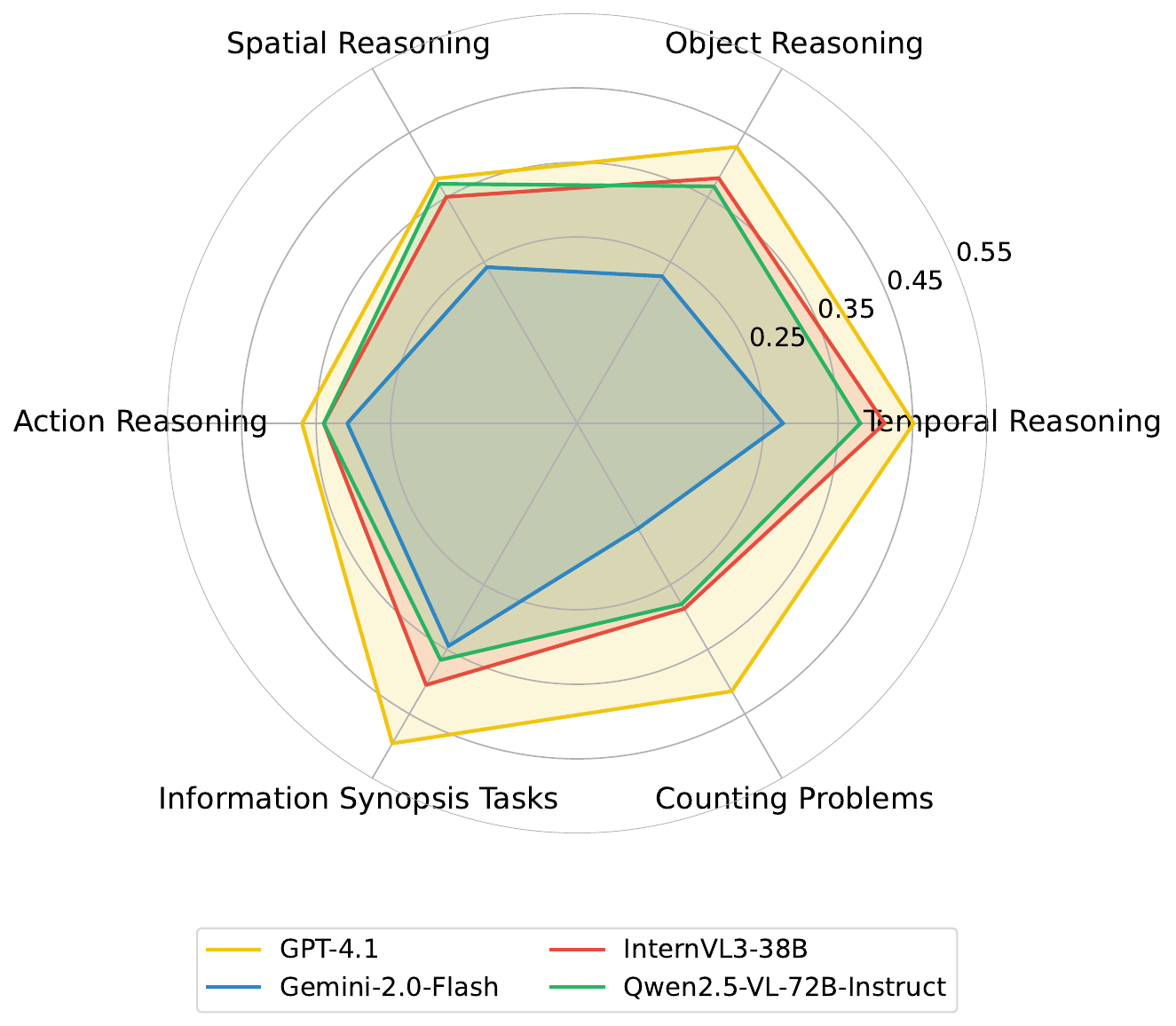}
    \caption{Performance comparison within four models on six reasoning sub-tasks. }
    \label{fig: sec5-reasoning}
\end{figure}

\subsection{Reasoning abilities analysis}
As mentioned in Section~\ref{sec:task-formulation}, we classify \RE task within six fine-grained reasoning abilities.  As illustrated in the \autoref{fig: sec5-reasoning}, GPT models and InternVL3-38B demonstrate stronger capabilities, particularly in tasks such as ``Information Synopsis'', ``Object Reasoning''. 
This may be attributed to their incorporation of more knowledge.
In contrast, models perform relatively worse on tasks like ``Spatial Reasoning'', ``Temporal Reasoning'', highlighting the challenges faced by these models in achieving competitive performance across various video understanding tasks. 

\subsection{Error Analysis}
\label{sec5-error-analysis}
To comprehensively evaluate the limitations of MLLMs, we perform detailed case studies and error analyses. The identified errors fall into these categories: 
(1) \textbf{Misconception of video creation}: This mistake is observed in \CV tasks. When asked to enhance a prompt, MLLMs often adhere closely to the initial prompt and give an expansion, highlighting a lack of understanding in video creation. 
(2) \textbf{Excessive Dependence on Textual Cues}: This issue is prevalent in \EA and \ED tasks, where MLLMs struggle with \multichoice questions requiring the identification of options violating physical laws. Faced with distractors that also breach such laws, MLLMs fail to differentiate and choose randomly.
(3) \textbf{Neglect of Critical Details}: This problem is evident across the four tasks proposed. MLLMs often miss crucial elements (\eg, blurriness, wind direction, camera shaking). 
(4) \textbf{Over-reliance on Commonsense Knowledge}: This typically occurs in \RE tasks. For instance, in ~\autoref{fig: overview}, when asked ``How many balls are on the table tennis table?'' some MLLMs might incorrectly answer ``Only one ball'' based on commonsense assumptions. 
Additional examples and illustrations are provided in Appendix \ref{appendix-error}.

\subsection{\method Analysis}
To assess the effectiveness of MLLM feedback in enhancing video generation, we examine whether human-in-the-loop feedback leads to improved results. As described earlier, annotators revise the LLM-generated prompts based on the content of the corresponding videos. These revised prompts are then used to re-generate the videos—a process we refer to as \method. We evaluate whether the newly generated videos exhibit higher quality compared to the originals.

The experiment is conducted on a dataset of 300 videos. Human judges are tasked with evaluating the quality of the re-generated videos in a pairwise comparison against the originals. For each video pair, annotators assess four aspects: subject consistency, background consistency, aesthetic appeal, and overall image quality. Appendix~\ref{reprompt-details} presents the detailed definitions of these evaluation criteria. 

\autoref{tab: sec5-reprompt} presents the win rates of the revised videos across these aspects. Overall, we find that videos generated from human-revised prompts consistently outperform the originals. Notably, improvements are most evident in subject consistency and aesthetic appeal. However, gains in visual quality and background consistency are more modest, likely reflecting inherent limitations of the underlying video generation model.
These results suggest that MLLMs have strong potential to enhance video generation, particularly if they can be more closely aligned with human preferences. When used as feedback providers or critics, MLLMs could drive meaningful improvements in generation quality. This perspective is further supported by recent work~\cite{lee2024videorepair}, which shows that integrating MLLM feedback into more sophisticated generation pipelines leads to notable gains in video quality.

\begin{table}[!t]
\centering
\small
\begin{tabular}{lr}
    \toprule
    Metrics & Win rate \\
    \midrule
    Subject  Consistency  & 56.7 \\
    Background  Consistency & 53.3 \\
    Aesthetic Quality  & 57.6  \\
    Image Quality & 50.7 \\
    \bottomrule
\end{tabular}
\renewcommand{\arraystretch}{1}  
\caption{Results (\%) of \method. 
}
\label{tab: sec5-reprompt}
\end{table}

\section{Conclusion}

Our experiments reveal that frontier MLLMs face significant challenges in \ours tasks involving \aivideos.
This poor performance is attributed to the unique characteristics of \aivideos. \aivideos often exhibit temporal inconsistencies, such as abrupt changes in motion or unnatural continuity between frames, and unrealistic object behaviors that defy commonsense. These challenges, combined with the semantic ambiguities and misalignment between visuals and textual input, make it difficult for MLLMs to generate feedback for video quality assessment. 
However, from our re-prompt pipeline, we observe that if MLLMs can better align with human preferences in video generation, their feedback becomes more valuable and trustworthy. Additionally, our experiments suggest that integrating other methods, such as computer vision techniques and in-context learning, can further enhance feedback precision. 
\section*{Limitations}
In this section, we outline three limitations of this study, each of which presents opportunities for future improvements. 
(1) First, only text-to-video models are considered, whereas videos generated from images may exhibit other types of error cases that are not addressed in this study. 
(2) Second, the design of the re-prompt pipeline is relatively simplistic, as it only incorporates textual feedback from humans. The specific positions of error cases are not included, which limits the granularity of the feedback. 
(3) Third, cross-modal videos are not included in our dataset. Since some video generation models also provide audio information, this omission may overlook more complex scenarios that arise from multimodal interactions.

\bibliography{anthology,custom, llm}

\appendix

\clearpage
\onecolumn
\section{\ours Annotation}
\label{appendix-annot}
\subsection{Annotator Biography}
The detailed biographies of the annotators involved in \ours construction are presented in ~\autoref{tab: appendix-annotator}. 
\begin{table*}[h]
\centering
\begin{tabular}{ccccc} 
\toprule[.1em]
\textbf{ID} & \textbf{Year} & \textbf{Major} & \textbf{\#Videos} & \textbf{Inter Agreement} \\
\midrule 
1 & Graduate or above & Geology & 162 & 78.4\% \\
2 & Fourth year  & Agricultural Resources and Environment & 200 & 78.9\%\\
3 & Third year  & Journalism & 136 & 91.8\%\\
4 & Fourth year  & Electrical Engineering and Automation & 154 & 90.3 \% \\
5 & Graduate or above & Mechanical Engineering and Automation & 162 &  86.1\%\\
6 & Graduate or above & Electrical Engineering & 160 & 88.5\% \\
7 & Graduate or above & Structural Engineering & 164 & 89.2\%\\
8 & Graduate or above & Electronics and Information & 279 & 76.9\%  \\
9 & Graduate or above & Clinical Medicine & 190 & 86.5\% \\
10 & Fourth year & Electrical Engineering and Automation & 182 & 91.3\% \\
11 & Graduate or above & Mechanical Engineering & 80 & 83.2\% \\
12 & Fourth year  & Polymer Materials and Engineering & 128 & 98.9\% \\
13 & Second year  & Cultural Heritage and Museum Studies & 165 & 93.7\% \\
14 & Second year  & Computer Science and Technology & 65 & 93.4\%  \\
15 & Graduate or above & Computer Science and Technology & 72 & 98.1\% \\
\bottomrule[.1em]
\end{tabular}
\caption{Annotator Details}
\label{tab: appendix-annotator}
\end{table*}

\subsection{Data Annotation and Validation}
The data annotation process primarily takes place on Label Studio, as mentioned above. To ensure the quality of the dataset, each video is reviewed by a separate reviewer. We also provide the annotator agreement, as shown in ~\autoref{tab: appendix-annotator}. For the annotators whose Annotator Inter Agreement score is less than 80\%, we require viewer to recheck his annotations. 

\clearpage
\subsection{Annotation Interface and Guideline  }
\begin{figure}[h]
    \centering
    \includegraphics[width=0.6\textwidth]{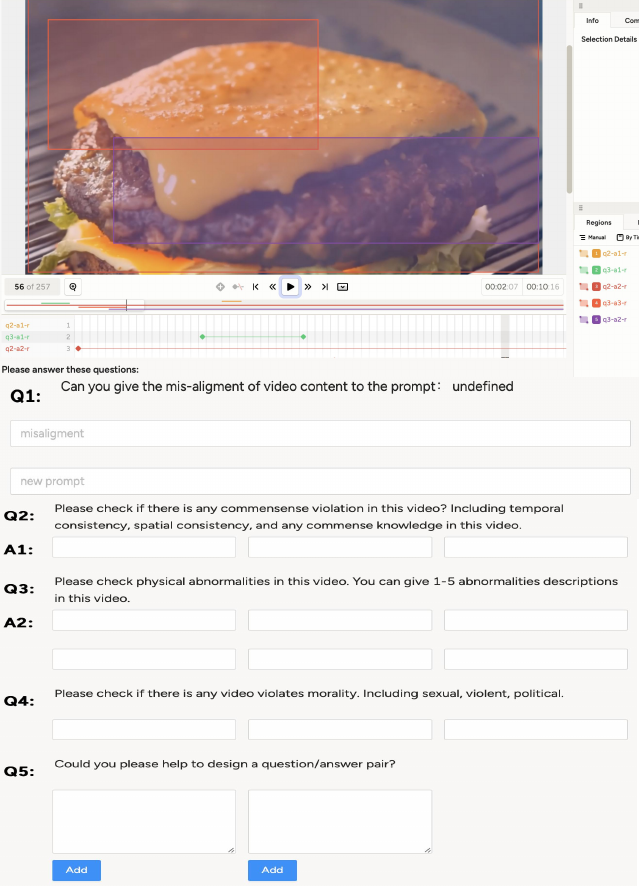}
    \caption{UI of annotation.}
    \label{fig: appendix-annot-ui}
\end{figure}

We employ Label Studio~\cite{LabelStudio} as our annotation platform. As shown in ~\autoref{fig: appendix-annot-ui}, we have five annotation tasks. 
Q1 is related to the \CV task, Q5 to the \RE task, and Q2–Q4 to the \EA and \ED tasks. For each annotator, they are first asked to complete a trial task, which involves annotating 10 videos. 
(1) For Q1, annotators are required to check for misalignment between the video and the text prompt. 
(2) For Q2–Q4, annotators are asked to select a specific error type. For example, for Q3, they can choose from categories such as "commonsense," "gravity," "lighting," and so on. 
Additionally, for Q2–Q4, annotators are asked to mark an area in the video that corresponds to the error type they have selected. 
(3) For Q5, human annotators are tasked with designing a question based on the specialties of \aivideos, and the question they design needs to require reasoning.

\clearpage
\section{Experiment Setup}

\subsection{MLLM Model Configuration}
\label{appendix-mllm-conf}
\autoref{tab: app-model-configuration} presents the details of model configuration for our experiments.
\begin{table*}[htbp]
\centering
\footnotesize
\resizebox{\textwidth}{!}{%
\begin{tabular}{llllcrrc}
\toprule
\textbf{Organization} & \textbf{Model} & \textbf{Release} & \textbf{Version} & \textbf{\begin{tabular}[c]{@{}c@{}}Support\\Video?\end{tabular}} & \textbf{\begin{tabular}[c]{@{}c@{}}\# Input\\Frames\end{tabular}} & \textbf{\begin{tabular}[c]{@{}c@{}}Inference\\Pipeline\end{tabular}} \\
\midrule
\multicolumn{8}{c}{\emph{\textbf{Proprietary 
 Models}}} \\
 \midrule
\multirow{2}{*}{OpenAI} & GPT-4.1 &  2025-4  &  \texttt{gpt-4.1-2025-04-14} &\xmark  &16 & \multirow{2}{*}{API}\\
& GPT-4.1-mini &  2025-4  &  \texttt{gpt-4.1-mini-2025-04-14} &\xmark &16 & \\
\noalign{\vskip 0.5ex}\hdashline\noalign{\vskip 0.5ex}

{Google}
 & Gemini-2.0-Flash & 2024-12 & \texttt{gemini-2.0-flash} &\cmark & 1fps &{API} \\
\midrule
\multicolumn{8}{c}{\emph{\textbf{Open-source Multimodal Foundation Models}}} \\
 \midrule
\multirow{2}{*}{LMMs-Lab} & LLaVA-NeXT-Video-7B & 2024-8 & \texttt{LLaVA-NeXT-Video-7B-hf} &\cmark & 16 & vLLM \\
& llama3-llava-next-8b-hf & 2024-7 & \texttt{llama3-llava-next-8b-hf} &\xmark &2 & vLLM \\
\noalign{\vskip 0.5ex}\hdashline\noalign{\vskip 0.5ex}

\multirow{1}{*}{Microsoft} & Phi-3.5-Vision & 2024-7 & \texttt{Phi-3.5-vision-instruct} &\xmark  &16 & vLLM\\

\noalign{\vskip 0.5ex}\hdashline\noalign{\vskip 0.5ex}

\multirow{2}{*}{Shanghai AI Lab} 

& InternVL3-8B & 2025-4 & \texttt{InternVL3-8B} &\xmark  &4 & vLLM\\
& InternVL3-38B & 2025-4 & \texttt{InternVL3-38B} &\xmark &4 & vLLM\\

\noalign{\vskip 0.5ex}\hdashline\noalign{\vskip 0.5ex}
\multirow{2}{*}{Alibaba} 

& Qwen2.5-VL-7B & 2024-9 & \texttt{Qwen2-VL-7B-Instruct} &\cmark  & 8 & vLLM\\
& Qwen2.5-VL-72B & 2024-9 & \texttt{Qwen2-VL-72B-Instruct} &\cmark  & 8 & vLLM\\

\noalign{\vskip 0.5ex}\hdashline\noalign{\vskip 0.5ex}
{DAMO} & VideoLLaMA3 & 2025-1 & \texttt{VideoLLaMA3-7B} &\cmark &1fps & HF \\

\noalign{\vskip 0.5ex}\hdashline\noalign{\vskip 0.5ex}
{Meta} & Llama-3.2-11B-Vision & 2024-9 & \texttt{Llama-3.2-11B-Vision-Instruct} &\xmark & 8 & {vLLM} \\
\noalign{\vskip 0.5ex}\hdashline\noalign{\vskip 0.5ex}
Mistral AI & Mistral-Small-3.1-24B & 2025-3 & \texttt{Mistral-Small-3.1-24B-Instruct-2503} &\xmark & 8 & {vLLM} \\
\bottomrule
\end{tabular}
}
\caption{ Details of the multimodal models in \ours. The ``Source'' column lists URLs for proprietary models and Hugging Face names for open-source models. The ``\# Input Frames'' column shows the default input frames, chosen from {2, 4, 8, 16}, based on the context window. ``HF'' refers to Hugging Face. }
\label{tab: app-model-configuration}
\end{table*}

\clearpage
\subsection{Prompts for Evaluation}
\label{appendix-prompt}
As mentioned, we primarily use the chain of thought (COT) technique to prompt MLLMs and obtain the \textbf{raw response}. The prompt for \yesno questions is shown in ~\autoref{fig: yn_cot_prompt}, and the prompt for \multichoice questions is shown in ~\autoref{fig: cot_prompt}. 
For \openend questions, we have two tasks: CV (\CV) and RE (\RE) to benchmark. The prompt for CV is shown in ~\autoref{fig: cv_shot_prompt}, and the prompt for RE is shown in ~\autoref{fig: op_shot_prompt}.  

As we benchmark a wide range of MLLMs, we do not expect that each model will output a well-formatted (\ie JSON) response. Thus, we use LLMs (\ie GPT-4o) to extract the answer from the \textbf{raw response} of those MLLMs. For \yesno and \multichoice questions, the LLM is used to directly extract the response; the corresponding prompts are shown in ~\autoref{fig: extract-yesno-prompt} and ~\autoref{fig: extract-multi-prompt}. For \openend questions, the LLM is required to compare the matching extent between the response of the MLLM and the correct answer. The prompt is shown in ~\autoref{fig: extract-openend-prompt}.

\begin{figure*}[htbp] 
\centering
\fbox{
    \begin{minipage}{\linewidth}      
        \textbf{[YN\_COT\_PROMPT]} \\
        Question: \{question\} \\
                Answer the given question. The last line of your response should be of the following format:  
                "Answer: {Your Answer}" ('yes' or 'no'), where ANSWER is the final answer of the questions. 
                Think step by step before answering. \\
    \end{minipage}%
}
\caption{Prompt for \yesno questions. }
\label{fig: yn_cot_prompt}
\end{figure*}

\begin{figure*}[htbp] 
\centering
\fbox{
    \begin{minipage}{\linewidth} 
        \textbf{[COT\_PROMPT]} \\
        Question: \{question\} \\
        Options: \{optionized\_str\} \\
                Solve the given multiple-choice question step by step. Begin by explaining your reasoning process clearly and thoroughly. After completing your analysis, conclude by stating the final answer using the following format: 'Therefore, the final answer is:  
        \{final\_answer\}. \\
    \end{minipage}%
}
\caption{Prompt for \multichoice questions. }
\label{fig: cot_prompt}
\end{figure*}

\begin{figure*}[htbp] 
\centering
\fbox{
    \begin{minipage}{\linewidth} 
        \textbf{[CV\_SHOT\_PROMPT]} \\
        Question: \{question\} \\
        Answer the given question. The last line of your response should be in the following format:  \\ 
        "Answer: {Your Answer}" (without quotes), where {Your Answer} is the final answer to the question.  \\ 
        Think step by step before answering. \\   
        Here are some example answers: \\ 
            1. "Basically in line with the situation, but other elements need to be added. The blue petal sunflower turns with the sun.", \\ 
            2. "Completely inconsistent with the text description. It should be changed to: One person holds a delicate smoke machine, which releases a large amount of gas in all directions." \\ 
            3. "The text seen through the glass does not match the original text. Look at the text through a glass of water." \\
    \end{minipage}%
}
\caption{Prompt for CV(\CV) tasks. }
\label{fig: cv_shot_prompt}
\end{figure*}

\begin{figure*}[htbp]
\centering
\fbox{
    \begin{minipage}{\linewidth}
        \textbf{[CV\_SHOT\_PROMPT]} \\
        Question: \{question\} \\
        Answer the given question. The last line of your response should be in the following format:  
        "Answer: {Your Answer}" (without quotes), where {Your Answer} is the final answer to the question.  
        If necessary, you can answer with phrases like 'not sure', 'violates reality', etc.  
        Think step by step before answering.  
        Here are some examples: \\
        1. Q: "Why does the amount of white sugar in the video increase as the spoon stirs?" \\
        A: "In real life, sugar does not increase with stirring a spoon; the content in the video goes against common sense." \\
        2. Q: "What color pants are the skaters wearing?" \\
        A: "Sometimes they are white, sometimes they are black." \\
        3. Q: "How many cars passed by?" \\
        A: "6-10 vehicles. At the beginning of the video, there is one vehicle, and later many vehicles flash in the frame, suggesting that this video might not depict a real-life scene."
    \end{minipage}
}
\caption{Prompt for RE(\RE) questions. }
\label{fig: op_shot_prompt}
\end{figure*}

\begin{figure*}[htbp]
\centering
\fbox{
    \begin{minipage}{\linewidth}
        \textbf{[EXTRACT\_YESNO\_RESPONSE\_PROMPT]]} \\
        Given a string, extract the reasoning process and the final answer from the string. \\  
        Output a JSON object with the following structure: \\ 
        { \\ 
          "reason": "The reasoning process derived from the string.", \\ 
          "answer": "yes"  \# The final answer either "yes" or "no". \\ 
        } \\ 
        Please don't include any other information in your response. \\  
        The string is {response}. \\
    \end{minipage}
}
\caption{Extraction prompt for \multichoice question. }
\label{fig: extract-yesno-prompt}
\end{figure*}

\begin{figure*}[htbp]
\centering
\fbox{
    \begin{minipage}{\linewidth}
        \textbf{[EXTRACT\_MULTICHOICE\_RESPONSE\_PROMPT]} \\
        Given a string, extract the reasoning process and the final answer from the string. \\ 
        Output a JSON object with the following structure: \\ 
        { \\ 
          "reason": "The reasoning process derived from the string.", \\ 
          "choices": "A"  \# The final answer, represented as an alphabet character (e.g., "A", "B", etc.). \\ 
        }
        Please don't include any other information in your response. \\  
        The string is {response}. \\ 
    \end{minipage}
}
\caption{Extraction prompt for \multichoice question. }
\label{fig: extract-multi-prompt}
\end{figure*}

\begin{figure*}[htbp]
\centering
\fbox{
    \begin{minipage}{\linewidth}
        \textbf{[EXTRACT\_OPENEND\_RESPONSE\_PROMPT]} \\
        Given the question, evaluate whether the response completely matches the correct answer. \\ 
        First, check the response and please rate score 0 if the response is not a valid answer. Please rate score 2 if the response completely or almost completely matches the correct answer on completeness, accuracy, and relevance. Please rate score 1 if the response partly matches the correct answer on completeness, accuracy, and relevance.  \\ 
        Please rate score 0 if the response doesn't match the correct answer on completeness, accuracy, and relevance at all. Please only provide the result in the following format: "Score:", \\ 
        No other information should be included in your response. \\ 
        
        Output a JSON object with the following structure: \\ 
        { \\ 
          "reason": "The reasoning process derived from the string.", \\ 
          "score": 0  \# The final answer, represented as an number (e.g., "0", "1", "2"). \\ 
        } \\ 
        Please don't include any other information in your response.  \\ 
                                      
        Question: {question} \\ 
        Response: {response} \\ 
        Correct Answer: {answer} \\
    \end{minipage}
}
\caption{Extraction prompt for \openend question. }
\label{fig: extract-openend-prompt}
\end{figure*}

\clearpage
\subsection{\method Details}
\label{reprompt-details}
We provide the definition of the metrics used in \method experiments here. 
(1) Subject Consistency: The degree to which the appearance, identity, or structure of the main subject remains stable across frames. 
(2) Background Consistency: The temporal coherence and spatial stability of the background throughout the video. 
(3) Aesthetic Quality: The overall visual appeal of the video based on artistic and stylistic elements. 
(4) Image Quality: The technical clarity and fidelity of each frame in the video.

And the final score is calculated as follows:
\begin{equation}
\text{Win Rate} = \frac{N_{\text{human}}}{N_{\text{total}}}
\label{eq:preference-score}
\end{equation}
\noindent
where \( N_{\text{human}} \) denotes the number of video pairs in which the human-generated video is preferred,  
and \( N_{\text{total}} \) is the total number of video pairs evaluated.
\clearpage
\onecolumn
\section{Case Study and Error Analysis}

\subsection{Examples of Reasoning Tasks}
\label{appendix-reasoning-examples}
\begin{figure*}[htbp]
    \centering
    \includegraphics[width=0.97\textwidth]{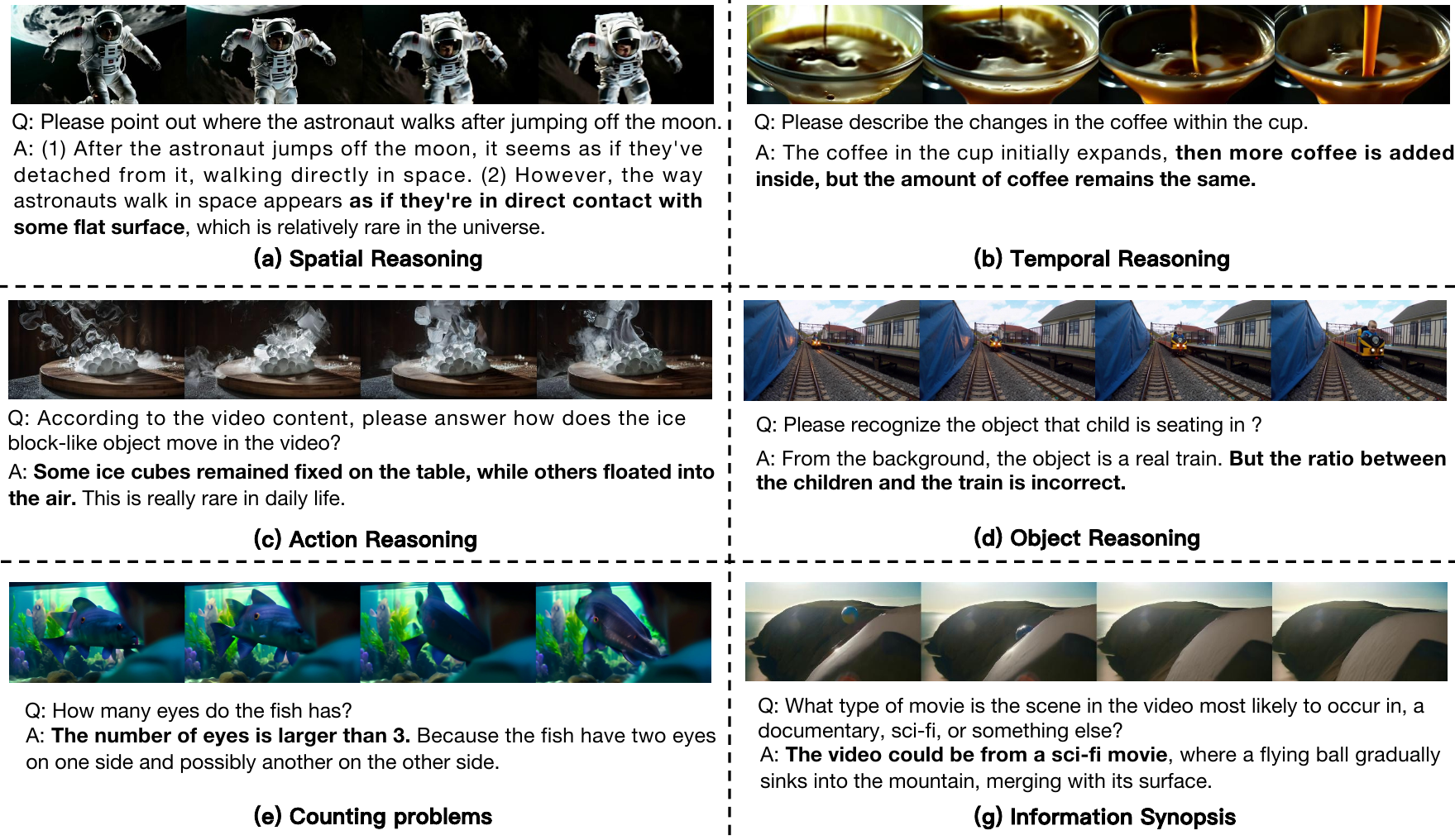}
    \caption{Detailed examples of reasoning tasks. }
    \label{fig: reasoning-examples}
\end{figure*}

We aim to benchmark the reasoning ability of MLLMs over six different reasoning tasks, including spatial and temporal reasoning, action and object reasoning, counting problems, and information synopsis.
We exemplify some scenarios of these tasks. 
(1) Spatial reasoning requires the MLLM to deduce the position or relative positioning of objects within the video. 
(2) In our dataset, temporal reasoning focuses on detecting anomalies that violate temporal correctness. (3) Action reasoning involves recognizing actions or distinguishing unusual actions based on objects or contextual cues. 
(4) Object reasoning primarily assesses the MLLM’s ability to identify object properties or correctly name objects. 
(5) For counting problems, the MLLM needs to detect the sudden appearance or disappearance of objects in videos. (6) Finally, information synopsis requires the MLLM to summarize the video's theme or infer its possible background. For clarification, we provide examples of these reasoning tasks as shown in ~\autoref{fig: reasoning-examples}.

\clearpage
\subsection{Error Analysis}
\label{appendix-error}
\begin{figure*}[htbp]
    \centering
    \includegraphics[width=0.95\textwidth]{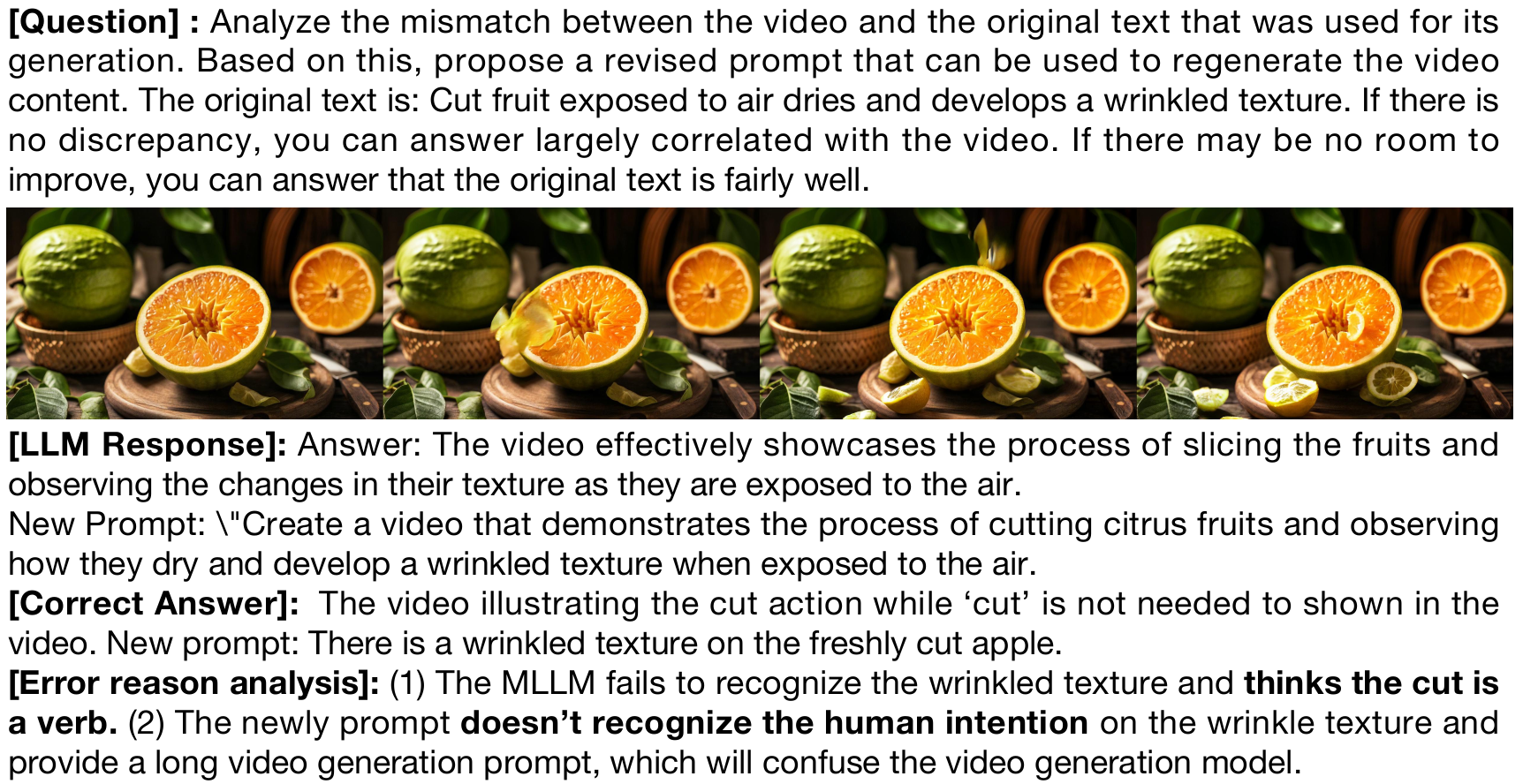}
    \caption{Misconception of Video Creation}
    \label{fig: error-misconceptions}
\end{figure*}

\begin{figure*}[htbp]
    \centering
    \includegraphics[width=0.95\textwidth]{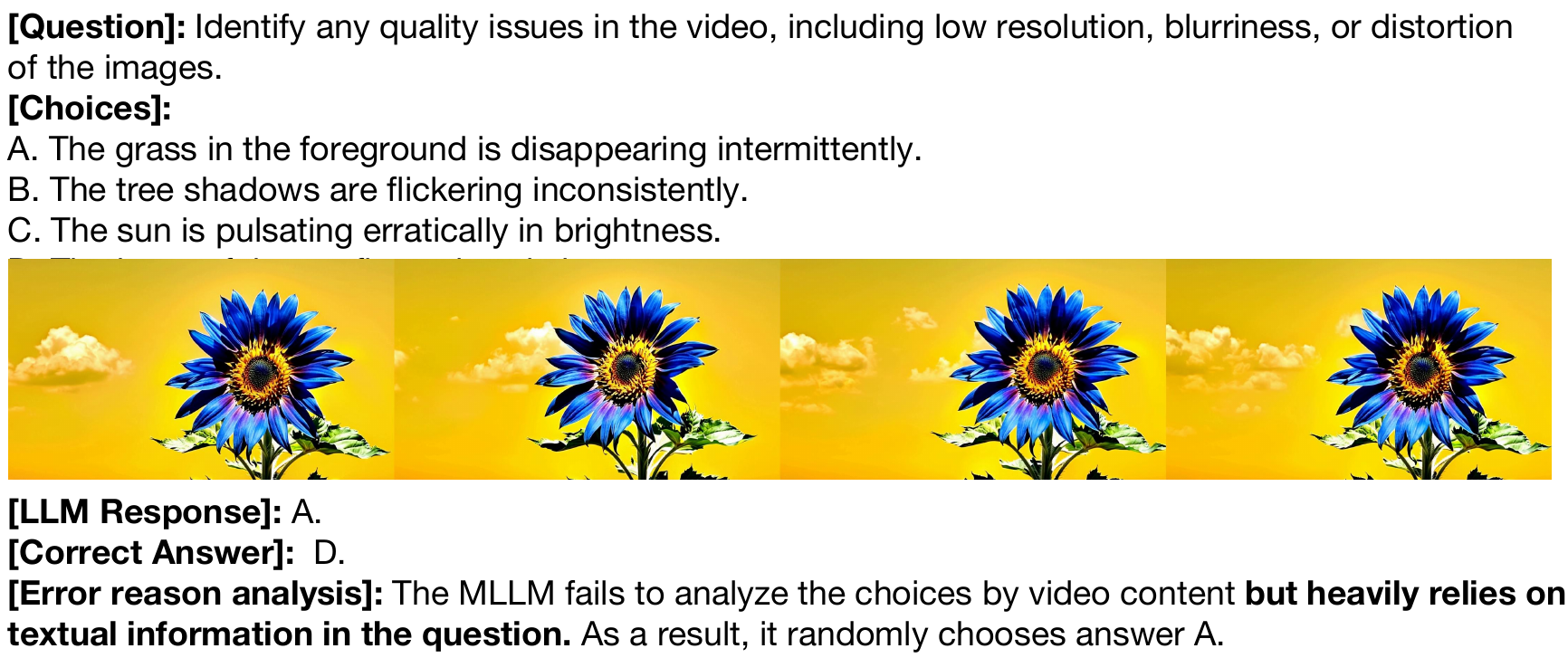}
    \caption{Excessive Dependence on Textual Cues}
    \label{fig: error-text-depend}
\end{figure*}

\begin{figure*}[htbp]
    \centering
    \includegraphics[width=0.95\textwidth]{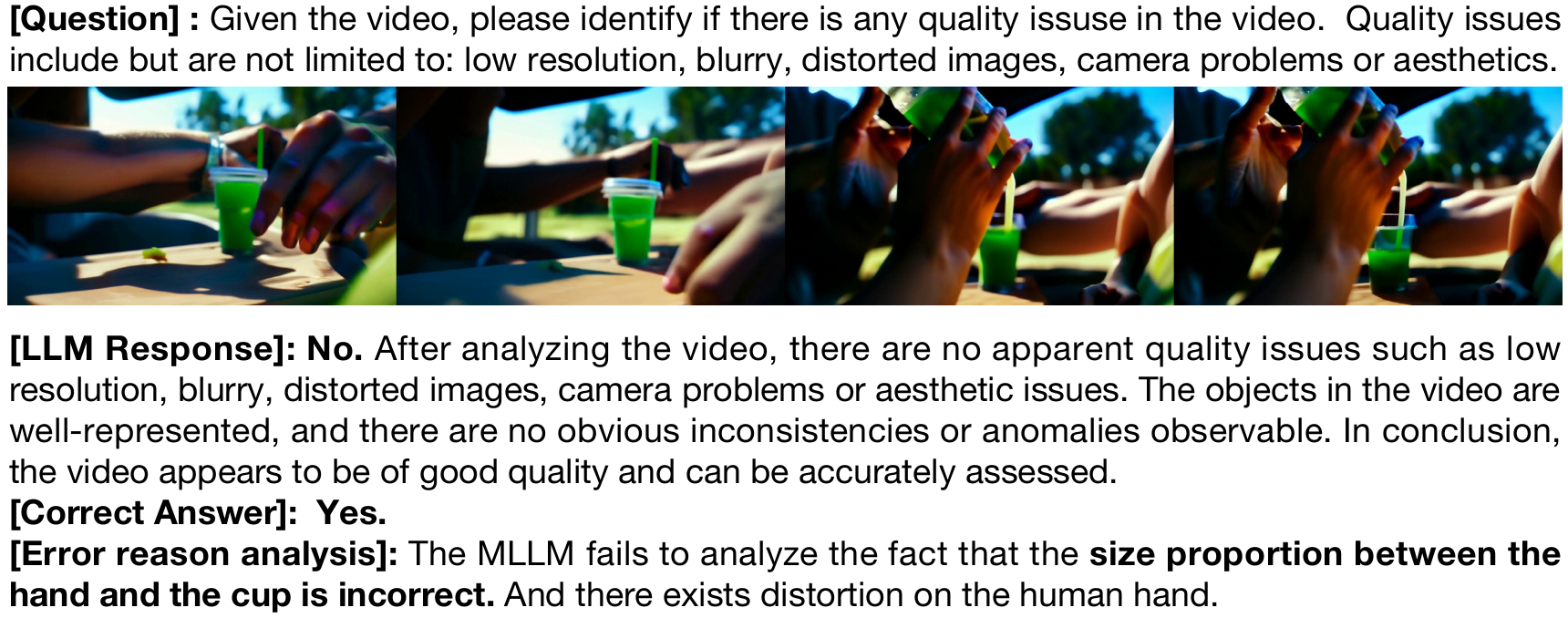}
    \caption{Neglect of Critical Details}
    \label{fig: error-neglect-details}
\end{figure*}

\begin{figure*}[htbp]
    \centering
    \includegraphics[width=0.95\textwidth]{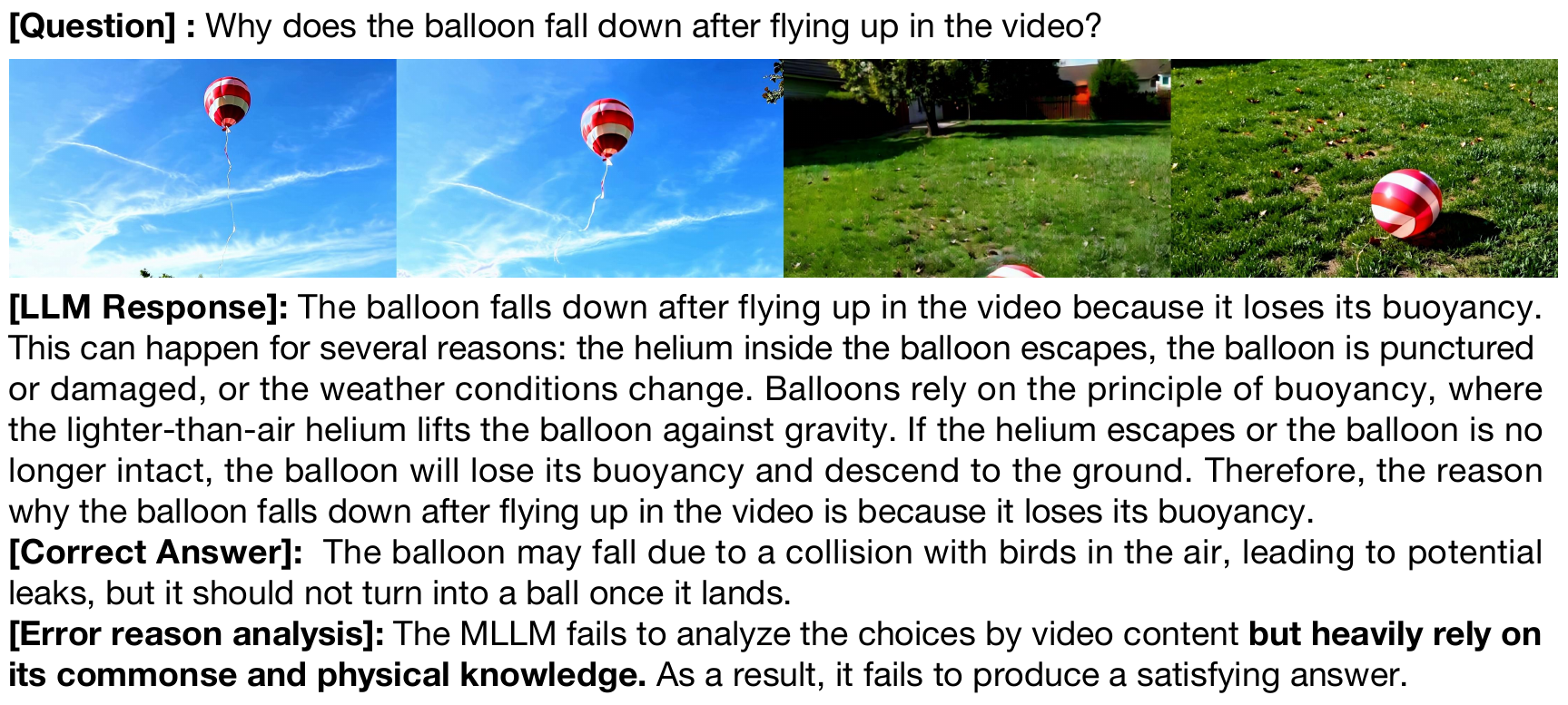}
    \caption{Over-reliance on Commonsense Knowledge}
    \label{fig: error-over-reliance}
\end{figure*}

As described in section ~\ref{sec5-error-analysis}, we have classify the errors of MLLM into four categories. 
Considering Qwen2-VL-72B-Instruct have a comparable performance against other models, we provide the error cases of it as follows:
(1) \textbf{Misconception of video creation}: ~\autoref{fig: error-misconceptions} shows that the MLLM doesn't understant how to generate a good video prompt. It may simply add more information and constraints to the original prompt. 
(2) \textbf{Excessive Dependence on Textual Cues}: ~\autoref{fig: error-text-depend} demonstrates that the MLLM may rely on its prior knowledge rather than observing details in the video.
(3) \textbf{Neglect of Critical Details}: ~\autoref{fig: error-neglect-details} shows that the MLLM tends to answer questions while overlooking critical details in \aivideos.
(4) \textbf{Over-reliance on Commonsense Knowledge}: ~\autoref{fig: error-over-reliance} illustrates the model's over-reliance on textual information.

\end{document}